\documentclass[acmsmall]{acmart} 

\usepackage{epsfig}
\usepackage{graphicx}
\usepackage{amsmath}
\usepackage{amssymb}
\usepackage{enumitem} 
\usepackage{subcaption}
\usepackage{textcomp}
\usepackage{hyperref}
\hypersetup{
    colorlinks=true,
    linkcolor=blue,
    filecolor=magenta,      
    urlcolor=cyan,
}
\usepackage[export]{adjustbox} 

\usepackage{mathtools}
\DeclarePairedDelimiter{\ceil}{\lceil}{\rceil}

\usepackage{colortbl}
\usepackage{arydshln,xcolor,array}

\AtBeginDocument{%
  \providecommand\BibTeX{{%
    \normalfont B\kern-0.5em{\scshape i\kern-0.25em b}\kern-0.8em\TeX}}}

\setcopyright{acmcopyright}
\copyrightyear{2020}
\acmYear{2020}
\acmDOI{10.1145/1122445.1122456}

\acmJournal{JACM}
\acmVolume{37}
\acmNumber{4}
\acmArticle{111}
\acmMonth{10}

\newcommand{\methodname}{DANTE} 

\acmSubmissionID{3838}

\begin{document}

\title{Improving Social Awareness Through DANTE: A Deep Affinity Network for Clustering Conversational Interactants}

\author{Mason Swofford}
\affiliation{%
  \institution{Stanford University} 
  \streetaddress{353 Serra Mall}
  \city{Stanford}
  \state{California}
  \postcode{94305}}
\authornote{Both authors contributed equally to this research.}
\email{mswoff@stanford.edu}
\author{John Peruzzi}
\affiliation{
\institution{Stanford University}
}
\authornotemark[1]
\email{jperuzzi@stanford.edu}

\author{Nathan Tsoi}
\affiliation{
\institution{Yale University}
  \streetaddress{51 Prospect St}
  \city{New Haven}
  \state{Connecticut}
  \postcode{06511}}
\email{nathan.tsoi@yale.edu}

\author{Sydney Thompson}
\affiliation{\institution{Yale University}}
\email{sydney.thompson@yale.edu}

\author{Roberto Mart\'in-Mart\'in}
\affiliation{\institution{Stanford University}}
\email{roberto.martinmartin@stanford.edu}

\author{Silvio Savarese}
\affiliation{\institution{Stanford University}}
\email{ssilvio@stanford.edu}

\author{Marynel V\'azquez}
\affiliation{\institution{Yale University}}
\email{marynel.vazquez@yale.edu}

\renewcommand{\shortauthors}{Swofford and Peruzzi, et al.}

\begin{abstract}
We propose a data-driven approach to  detect conversational groups by identifying spatial arrangements typical of these focused social encounters. Our approach uses a novel Deep Affinity Network (\methodname) to predict the likelihood that two individuals in a scene are part of the same conversational group, considering their social context. The predicted pair-wise affinities are then used in a graph clustering framework to identify both small (e.g., dyads) and large groups. The results from our evaluation on multiple, established benchmarks suggest that combining powerful deep learning methods with classical clustering techniques can improve the detection of conversational groups in comparison to prior approaches. Finally, we demonstrate the practicality of our approach in a human-robot interaction scenario. Our efforts show that our work advances group detection not only in theory, but also in practice.
\end{abstract}




\maketitle

\section{Introduction}
Automatic detection of group conversations enables a rich set of intelligent, social computer interfaces. For example, group detection has traditionally enabled surveillance systems~\cite{hung2011detecting, cristani2013human, setti2015f}, socially-aware mobile systems \cite{marquardt2012cross}, interactive displays \cite{ichino2016effects}, exhibits \cite{dim2015automatic}, and social playgrounds \cite{moreno2013socially, jungmann2014spatial}.  
In the context of robotics, group detection is also essential for situated spoken language interaction~\cite{bohus2014directions}, non-verbal robot behavior generation~\cite{vazquez2017towards}, and socially-aware robot navigation in human environments~\cite{rios2015proxemics}. However, detecting conversations in dynamic human environments is an intricate problem, which requires perceiving subtle aspects of social interactions. 

In this work, we study the problem of visually recognizing situated group conversations by analyzing \textit{proxemics} --- people's use of physical space \cite{hall1910hidden}. In particular, we study the automatic recognition of spatial patterns of human behavior that naturally emerge during group conversations \cite{goffman2008behavior}. These patterns are known as Face Formations, or \textit{F-Formations} in short, as denoted by A. Kendon~\cite{kendon1990}. They are the result of people needing to communicate in close proximity while sustaining a shared, focus of attention. Prototypical formations are often observed as face-to-face, side-by-side or circular spatial arrangements in open spaces. However, the specific type of arrangement that emerges during conversations ultimately depends on a number of social factors, including the number of interactants, their conversation topic,  and environmental spatial constraints. 
F-Formations provide a conceptual framework in Human-Computer Interaction (HCI) for thinking about how the physical aspects of a setting influence interactions \cite{ballendat2010proxemic, marshall2011using, de2014reorganizing, tong2016s}.



Most prior work on visual F-Formation detection  has focused on explicitly modeling properties of conversational group spatial arrangements~\cite{hung2011detecting, setti2013multi, setti2015f}. For instance, people tend to keep a social distance from one another during conversations~\cite{hall1910hidden} and orient their bodies towards the center of their group~\cite{kendon1990}. But these approaches do not typically account for the malleability inherent in human spatial behavior. For example, people naturally adapt to crowded environments and modify their spatial formations by interacting closer if need be. 
Robustness to these complex scenarios is essential for reasoning about group conversations through spatial analysis in real applications.

\begin{figure}[t]
    \centering
    \includegraphics[width=\linewidth]{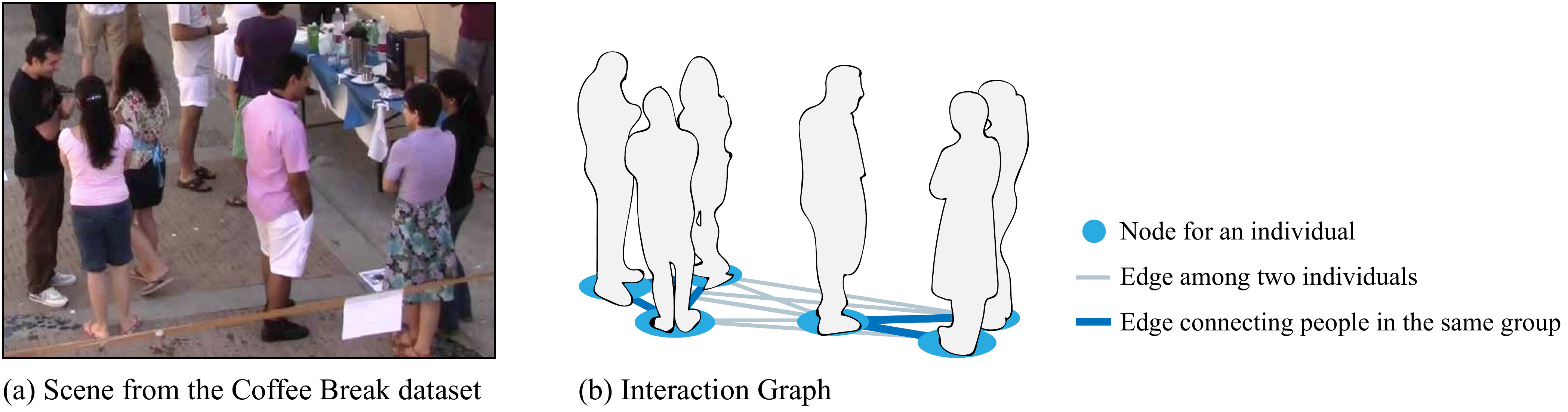}
    \caption{Example problem setting. Given a social scene (a), the goal is to identify the individuals that are part of the same conversational group based on their position and orientation relative to one another (b). We approach this problem by combining graph clustering with deep learning. See the text for more details.}
    \label{fig:pull}
\end{figure}


In contrast to most prior work on F-Formation detection, we explore using the powerful approximation capabilities of Deep Learning (DL) for identifying conversations and their members, but how can one leverage these approximation capabilities effectively given the small-sized datasets that are available for F-Formation detection? Additionally, how can one deal with variable number of group interactants? To answer these questions, we revisit ideas from  classical graph clustering solutions~\cite{yu2009monitoring, hung2011detecting, vascon2016detecting}. We view the problem of F-Formation detection as finding sets of related nodes in an \textit{interaction graph} (Figure \ref{fig:pull}). The nodes of the graph correspond to individuals in a scene with associated spatial features obtained through image processing. The graph edges connect two nearby people and have an associated affinity (weight) that encodes the likelihood that they are conversing. Under this framing, the key challenge for F-Formation detection is to compute appropriate affinities for identifying groups. While prior work  used simple heuristics to compute edge weights~\cite{yu2009monitoring, hung2011detecting, vascon2016detecting}, we propose to learn a function that predicts these weights.

Our main contributions are threefold:
\begin{enumerate}[leftmargin=*]
\item We propose a novel \textbf{D}eep \textbf{A}ffinity \textbf{N}e\textbf{T}work for clust\textbf{E}ring conversational interactants (\methodname). The network approximates the likelihood that two individuals are conversing given (a) their spatial features and (b) information about other nearby individuals, i.e., their social context. The network can deal with contexts with varying number of people by leveraging recent ideas to input mathematical sets to DL  \cite{qi2017pointnet}. 
\item We conduct an extensive evaluation of \methodname\ in established benchmarks. The evaluation shows that our proposed approach advances conversational group detection in comparison to state-of-the-art methods. 
Furthermore, our results show that DANTE is extensible. It can easily reason about different relevant spatial features for group detection. 
\item Finally, we demonstrate the applicability of \methodname\ on a social robotic system. In this context, \methodname\ enables a robot to identify the members of its  group conversations as well as nearby bystanders, who offer opportunities for new interactions. We open-source our code to facilitate future replication efforts and enable the use of \methodname\ to create other socially-aware interfaces.
\end{enumerate}

\section{Related Work}
A variety of approaches have been proposed to computationally detect situated group interactions through proxemics analysis within HCI \cite{brdiczka2005automatic,choudhury2002sociometer,hung2014detecting,marquardt2012cross}, computer vision \cite{groh2010detecting,fathi2012social, park20123d, alletto2014ego}, social signal processing \cite{olguin2009sensible, chen2011discovering}, and even natural language processing \cite{wyatt2007conversation}. Due to limited space, this section focuses on describing close related work on F-formation detection based on visual spatial features. For a broader review, we encourage interested readers to refer to \cite{aggarwal2011human} and \cite{Vazquez-2017-thesis} (Section 3.2).

\textbf{Visual conversational group detection.} Our work focuses on the analysis of visual human spatial behavior because: (1) visual sensing with cameras is cheap and does not require instrumentation of users, thus enabling group detection in unconstrained settings \cite{bohus2009learning}; (2) relevant visual features are readily available through open-source or commercial software \cite{cao2018openpose,kinect_azure}, (3) prior work has shown that these visual features are effective to enable proxemics interactions \cite{marquardt2011proximity}. It is worth noting that early research on conversational group detection within the computer vision community was motivated by surveillance applications in public human environments \cite{yu2009monitoring, ge2009automatically, choi2009they, chang2011probabilistic}. These approaches identified two key features for spatial analysis: human \textit{position} and \textit{orientation} information. We also use these features in our work.

Most prior approaches for visually detecting F-Formations are based on mathematical models of sustained spatial arrangements \cite{cristani2011social, setti2013multi, gan2013temporal, setti2015f, vazquez2015parallel}. These model-based approaches tend to formalize the \textit{transactional segments} of individuals, which is the space that extends forward from their lower body and that includes whatever they are currently engaged with. Then, these methods find the intersection of transactional spaces in a scene, or \textit{o-spaces} of the F-Formations \cite{kendon1990}. For example, \cite{cristani2011social,setti2013multi,vazquez2015parallel} use voting schemes to find o-spaces, while Setti et al. \cite{setti2015f} use an iterative graph-cuts approach. We consider the latter work \cite{setti2015f} in our evaluation because it provides state-of-the-art results on group detection using spatial features from a single image. 

\textbf{Group detection as graph clustering.} 
An important category of group detection methods rely on graph clustering  \cite{yu2009monitoring, hung2011detecting, vascon2016detecting}, including ours. In this setting, individuals correspond to nodes in a weighted, interaction graph (Fig. \ref{fig:approach}b). The goal is to partition the graph into groups of nodes that represent human interactions. Note that soft group assignments are also possible \cite{chang2011probabilistic,vazquez2015parallel}, but we focus on the hard assignment problem in this work because standard evaluation benchmarks provide hard group labels. 

Similar to \cite{hung2011detecting, vascon2016detecting}, we use the Dominant Sets  clustering algorithm to detect F-Formations in social scenes. Different to these prior efforts, though, we do not use hand-crafted heuristics 
\cite{hung2011detecting} nor a model of human attention \cite{vascon2016detecting} to assign weights to graph edges and perform clustering. Instead, we propose to learn in a data-driven fashion a non-linear function that predicts the weights. 
This idea is in line with   \cite{hedayati2019recognizing}, where it is proposed to to detect F-Formations by 
leveraging machine learning. However, instead of estimating binary  affinities between interactants and then finding cliques through a voting scheme  \cite{hedayati2019recognizing}, we estimate a continuous likelihood for people interacting together. One of our insights in this prediction problem is to consider the spatial information of individuals nearby the corresponding dyad of interest for a graph edge, i.e., to consider the dyad's \textit{social context}.

\textbf{Deep learning (DL) applied to group detection.}
Two recent methods have attempted to use DL for conversational group detection by reasoning about spatial information. First, \cite{arxivearlier} leveraged DL for estimating the location of o-spaces in a scene. Then, this prior method used a greedy geometrical approach to assign interactants to group conversations. We omit evaluating our approach against \cite{arxivearlier} since their results were far below the state-of-the-art \cite{setti2015f}. Second, Sanghvi, Yonetani and Kitani \cite{Sanghvi2018-rssw} proposed to use DL in the context of learning communication policies. As part of their policy network, the authors introduce a communication gating module that automatically infers
group membership. We consider this approach as a baseline in our evaluation given that their results are comparable to \cite{setti2015f}. 

\textbf{Group detection evaluation.}
We evaluate our approach on standard datasets for group detection within the computer vision and social signal processing community \cite{zen2010space, cristani2011social, FriendsMeet, SALSA}. 
The datasets provide ground truth group annotations, as well as spatial features  
for the individuals in the scene.  The latter features were gathered with automated computer vision techniques, thus providing realistic inputs for our experimental evaluation.\footnote{Even though A. Kendon \cite{kendon1990} defined transactional segments based on people's lower body orientation \cite{kendon1990}, we often use head orientation as a key feature for group detection. Our rationale for this decision is evaluating our approach using established  datasets~ \cite{cristani2011social, zen2010space}. As in \cite{ricci2015uncovering}, though, we believe that future work should consider both body and head orientation for interaction analysis. Our method could easily be extended to this end.}

Several prior efforts have demonstrated the value of F-formation detection in HCI. For example, F-formation detection has enabled spatial interfaces \cite{ballendat2010proxemic} and the generation of coordinated, non-verbal robot behavior in situated human-robot conversations  \cite{vazquez2017towards}. Likewise, we  demonstrate the applicability of our proposed approach in a real interactive system. This effort reinforces the value of F-Formation detection for building socially-aware interactive systems. 

\section{Conversational Group Detection with DANTE}

\begin{figure*}[t]
    \centering
    \includegraphics[width=\linewidth]{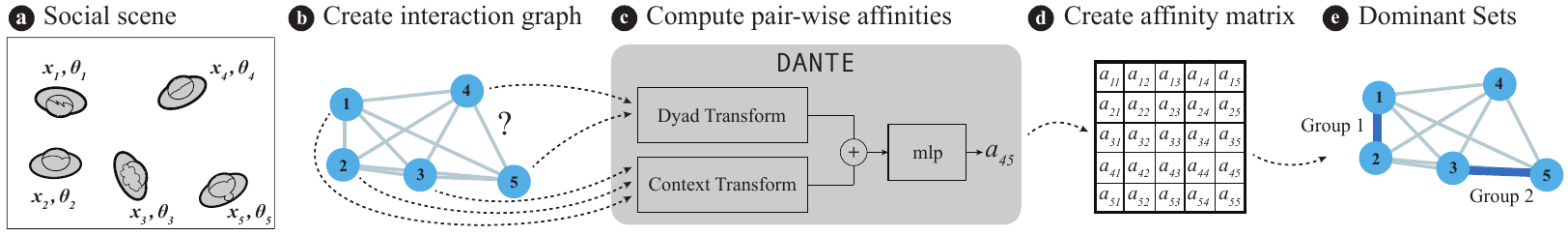}
    \caption{Group detection approach. Our method receives as input spatial features (e.g., position $\mathbf{x}$ and orientation $\theta$) for the social agents in a scene (a). This information is used to create an interaction graph (b) and to compute pair-wise affinities with DANTE (c). The affinities are assembled into an affinity matrix (d) to cluster nodes  (e). Multi-layer perceptron is abbreviated as mlp in (c). See the text for more details.}
    \label{fig:approach}
\end{figure*}

Our group detection method is illustrated in Fig.~\ref{fig:approach}. The input to our method is a set of $N$ potential interactants in a scene $\mathcal{I} = \{i_0,\ldots,i_N\}$. These interactants are typically people but could also be other social agents relevant for spatial analysis, like robots \cite{huttenrauch2006investigating,kuzuoka2010reconfiguring,vazquez2017towards}. Each $i_i= (\textit{id}_i, \mathbf{f}_i)$ in $\mathcal{I}$ has a unique identifier $\textit{id}_i$ for the social agent and a feature vector $\mathbf{f}_i$ with corresponding spatial information. By default, we include in the vector $\mathbf{f}_i$  the 2D position of the individual in the planar layout of the environment, $\mathbf{x}_i = (x_i,y_i)$, and its orientation, $\theta_i$, relative to a world coordinate frame (Fig.~\ref{fig:approach}(a)). Our rationale for including these features stems from prior work both in social psychology \cite{goffman2008behavior,kendon1990} and computer science \cite{chang2011probabilistic,cristani2011social,vascon2016detecting} that have shown the importance of these features for modeling F-Formations. But more generally, $\mathbf{f}_i$ could also include other spatial information, like the interactant's instantaneous velocity, as later discussed in our evaluation. 
%
%
The output of our method is another set, but in this case of $K$ detected conversational groups, $\mathcal{G}=\{g_0,\ldots,g_K\}$. Each  group $g_k$ is composed of the identifiers of the interactants that belong to it, $g_k = \{\textit{id}_i\}$. The conversational groups are mutually exclusive: a social agent can only belong to a single conversational group.

\subsection{Approach}
Our approach represents the scene as an  interaction graph $G = (V, E, A)$ with a set of nodes or vertices $V$, edges $E$, and non-negative affinities or  weights $A$. As shown in Fig.~\ref{fig:approach}(b), each node corresponds to a social agent and contains its data $i_i= (\textit{id}_i, \mathbf{f}_i)$. Pairs of nodes are connected by undirected edges in the graph. For each edge, its affinity score is meant to represent the likelihood that the two agents connected to the edge belong to the same group conversation. 

The main insight of our work is learning the  graph affinities such that we can  effectively partition the graph to determine the groups $\mathcal{G}$. In particular, we propose 
DANTE to predict affinity scores (Fig.~\ref{fig:approach}(c)). The benefits of DANTE include reducing the reliance of group detection on heuristics in comparison to prior work, e.g.,  \cite{yu2009monitoring, hung2011detecting, vascon2016detecting, setti2015f, vazquez2015parallel}. Section \ref{sec:dante_detail} further explains DANTE in detail.

Our proposed F-Formation detection approach finally builds an affinity (or similarity) matrix to cluster nodes with the Dominant Sets (DS) algorithm~\cite{hung2011detecting} (Fig.~\ref{fig:approach}(d)-(e)). The DS algorithm iteratively finds clusters that describe compact structures,
which are well suited to represent F-Formations. More details on this graph clustering procedure are provided in Sec. \ref{ssec:dominant_sets}.





\subsection{Affinity Scoring with \methodname}
\label{sec:dante_detail}



This section introduces DANTE, a neural network that predicts the weights for each of the edges in the social graph (Fig. \ref{fig:approach}c). 
DANTE is structured to reason about two types of information: local spatial information from the two vertices (individuals) connected to an edge of interest, and global spatial information from other nearby people, i.e., the social context of the dyad of interest. Predicting affinities based on these two types of information within a graph clustering framework is a novel contribution of our work. Because of it, our approach does not need additional ad-hoc steps \cite{setti2015f, vazquez2015parallel} to verify that the detected groups effectively conform with the notion of F-Formations~\cite{kendon1990}. DANTE's  structure and its inherent data-driven nature make it easily  extensible to applications with a variety of spatial features, as demonstrated in our evaluation.  

Without loss of generality, assume for the following sections that DANTE is computing the affinity $a_{ij}$ for the individuals $i$ and $j$ in the interaction graph $G$.

\subsubsection{Dyad Transform}
\begin{figure}[t]
    \centering
    \includegraphics[width=.55\linewidth]{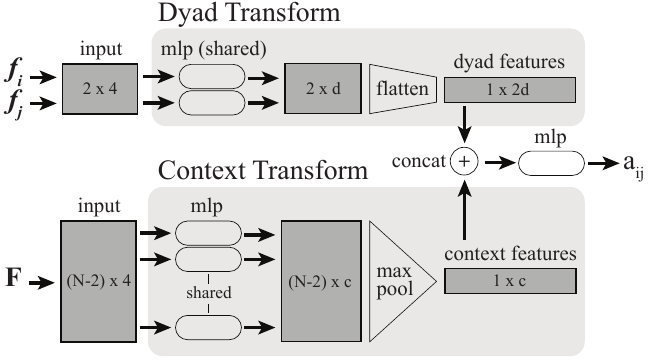}
    \caption{DANTE components. The pairwise affinity $a_{ij}$ of a pair of individuals $i$ and $j$ is computed from two types of features: the local \emph{dyad features} and the global \emph{context features}. 
    As explained in Sec. \ref{sec:implementation}, all spatial data that is input to DANTE is transformed to the canonical frame between $i$ and $j$ before computation. The abbreviations "concat" and "mlp" stand for concatenation and multi-layer perceptron, respectively.
    }
    \label{fig:dante}
\end{figure}

The Dyad Transform of DANTE is in charge of computing \textit{local features} for  $i$ and $j$, as depicted in the top part of Fig.  \ref{fig:dante}. The input to the Dyad Transform is a matrix with two rows, one for the feature  $\mathbf{f}_i$ and one for $\mathbf{f}_j$.
%
%
Each of these features are transformed independently by a multi-layer perceptron (mlp), resulting in a feature encoding for each social agent of dimensionality $d_{\textit{dyad}}$.
The mlp is composed of $D_{\textit{dyad}}$ dense layers followed by ReLU activations. The result is a matrix of features in $\mathbb{R}^{2 \times d_{\textit{dyad}}}$. This matrix is finally flattened into a dyad feature vector $\mathbf{v}_{\textit{dyad}} \in \mathbb{R}^{1 \times 2d_{\textit{dyad}}}$.

\subsubsection{Context Transform}

DANTE's Context Transform computes a \textit{global feature} representation for the social context of the dyad of interest, as illustrated in the bottom part of Fig.  \ref{fig:dante}. Our design for this model component is inspired by prior work on inputting sets to neural networks, especially in the context of point cloud processing~\cite{qi2017pointnet}. Our experimental results reinforce the notion that neural networks can encode spatial features in a scalable  manner for social interaction analysis~\cite{gupta2018social, pokle2019deep}. 


At its core, the Context Transform uses a symmetric function to handle unordered and potentially variable number of inputs. In our case, these inputs correspond to the set $\mathbf{F}$ of individual, spatial feature vectors 
for the people in the scene other than the dyad of interest. More formally, assume again that DANTE is computing the affinity $a_{ij}$. Then, the input to the Context Transform is a feature set $\mathbf{F} = \{ \mathbf{f}_k\,|\, 1 \leq k \leq N \ \text{and}\ k \not\in \{i,j\} \}$ with each vector $\mathbf{f}_k$ encoding the agent's $k$ spatial information (e.g., position and orientation) 
as explained in previous sections. The set $\mathbf{F}$ is implemented as a matrix with one feature per row.

Similarly to the Dyad Transform, the Context Transform first applies independently a multi-layer perceptron to each of the rows of its input matrix $\mathbf{F}$ (bottom part of Fig.~\ref{fig:dante}). The mlp is composed of $D_{\textit{context}}$ dense layers followed by ReLU activations, resulting in a matrix of features in $\mathbb{R}^{(N-2) \times d_{\textit{context}}}$. The latter matrix is finally transformed by max pooling along its rows. The output is a context feature vector $\mathbf{v}_{\textit{context}} \in \mathbb{R}^{1 \times d_{\textit{context}}}$. Note that max pooling is the key symmetric operation that makes the Context Transform invariant to input permutations.

\subsubsection{Combining Dyad and Context Features}

Finally, DANTE combines the dyad information $\mathbf{v}_{\textit{dyad}}$ and context information $\mathbf{v}_{\textit{context}}$ to compute the affinity score $a_{ij}$. To this end, DANTE first concatenates the two feature vectors column-wise, resulting in a new vector in $\mathbb{R}^{1\times (2d_{\textit{dyad}}+d_{\textit{context}})}$. Then, an mlp is used to transform the combined features into a joint representation. In this case, the mlp is composed of $D_{\textit{comb}}$ dense layers, each followed by ReLU activations. Finally, one more dense layer projects down the resulting features into a scalar value. This last layer uses a sigmoid activation function to constraint the output to the $[0,1]$ range.

\subsubsection{Other Implementation Details}
\label{sec:implementation}
\begin{figure}[t!p]
    \centering
    \includegraphics[width=.3\linewidth]{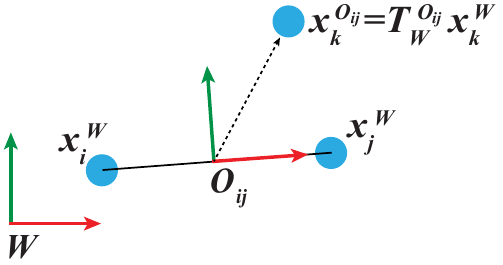}
    \caption{Localizing transformation. When computing a given affinity $a_{ij}$, we transform the features input to DANTE from the world frame $W$ to a local frame $O_{ij}$ relative to the two potential interactants of interest ($i$, $j$).}
    \label{fig:frame}
\end{figure}

The spatial features $\mathbf{f}_i$ are originally obtained in a world reference frame $W$. However, we transform them before inputting them to DANTE to a canonical frame of reference defined with respect to the pair of individuals whose affinity $a_{ij}$ is being computed. This canonical frame $O_{ij}$ is illustrated in Fig. \ref{fig:frame}. 
The frame $O_{ij}$ is located at the middle point between the social agents $(\mathbf{x}_i + \mathbf{x}_j)/2$ in the global frame $W$. For setting orientation of $O_{ij}$, we align its $x$ axis with a vector from the position of $i$ to the position of $j$. 
This transformation facilitates learning and generalization. 

\noindent


By default, we use 4-dimensional representation for the spatial features $\mathbf{f}_i$ relative to $O_{ij}$ in our implementation (Fig. \ref{fig:dante}). The four dimensions correspond to the 2D position ($\mathbf{x}_i$) and the orientation $\theta_i$ of the social agent, with the angle encoded through $\sin(\theta_i)$ and $\cos(\theta_i)$. Using sine and cosine helps avoid issues with $\theta_i$ wrapping around $360^\circ$. Another benefit is that the projection forward in the direction that an agent is looking is the result of a simple multiplication of the sine and cosine of the orientation by some positive value. This projection has been employed by several other group detection algorithms \cite{cristani2011social, setti2013multi, setti2015f}, suggesting that it can facilitate reasoning about F-Formations. We theorize that the proposed representation can be used by \methodname\ to easily learn to process spatial data in a useful manner. Furthermore, we show in Sec. \ref{sec:generalization_experiment} how \methodname\  can be adapted to alternative spatial feature  representations when orientation information is not readily available.


\subsection{Dominant Sets Grouping}
\label{ssec:dominant_sets}

Once the affinities for each pair of individuals in the social interaction graph are computed, our approach proceeds to group people using the Dominant Sets (DS) algorithm by Hung and Kr{\"o}se~\cite{hung2011detecting}. Dominant sets (clusters) in the algorithm are a generalization
of maximal cliques to edge-weighted graphs with no self-loops~\cite{pavan2007dominant}. Our social interaction graph $G$ is one such graph. 


For a detailed explanation of the Dominant Sets algorithm, we refer interested readers to Sections 3.2 and 6 of \cite{hung2011detecting}. In short and for completeness, the DS algorithm iteratively finds clusters that satisfy the following property: the mutual affinity, formally defined in \cite{hung2011detecting}, between all of the cluster members is higher than the affinity between  its members and those outside of it. Once a cluster is found, new clusters that satisfy the same property are then searched for. 
The stopping criterion for finding clusters in the DS algorithm is reached when (a) a new cluster either does not satisfy the property of high relative mutual group affinity or (b) when the mutual affinity of a group is below a certain threshold, which is  determined through cross validation. In general, the DS algorithm results in  compact clusters that are well suited to represent F-Formations of any size.

Although DS can be applied to social interaction graphs with asymmetric affinities, symmetric affinities have been reported to yield superior results~\cite{hung2011detecting,vascon2016detecting}. Thus, we assume symmetric affinities in our work by setting edge weights to the average of the predicted $a_{ij}$ and $a_{ji}$, for $1 \leq i, j \leq N$ and $i \not = j$. Note that we could have made DANTE directly output symmetric affinities by making the Dyad Transform permutation invariant, like the Context Transform. However, we found in our initial tests that this design led to reduced performance in comparison to averaging $a_{ij}$ and $a_{ji}$. We hypothesize this is because our non-symmetric architecture can better reason about the Dyad's spacial features, compared to the symmetric architecture, which cannot leverage the ordering of the Dyad inputs.

\section{Evaluation}
\label{sec:experiments}

We conduct systematic evaluations of our proposed group detection approach using established benchmarks. In this section in particular, we first describe experiments with datasets that were created specifically for the evaluation of conversational group detection algorithms. These datasets provide position and orientation information for the individuals in a scene, as needed for our default spatial feature representation (Sec. \ref{sec:implementation}). In Section \ref{sec:generalization_experiment}, we further evaluate \methodname\ in a different setup to test its generalization capabilities to the more general problem of group detection. This involves the detection of F-Formations, but potentially also other types of group formations. 
Worth noting, the Appendix provides an additional experiment with synthetic data, which is not discussed outside Sec. \ref{Synthetic Data Augmentation}  because results were inconclusive.


\subsection{Datasets}

We consider three publicly available datasets of social interactions in our evaluation, presented below in order of   annotation quality:
\begin{itemize}[leftmargin=*]
    \item[--] \textit{Cocktail Party Dataset} \cite{zen2010space}. Contains about 30 min. of video recordings of a cocktail party in a lab environment. The video shows 6 people conversing with one another and consuming drinks and appetizers. The party was recorded using four synchronized cameras installed in the corners of the room. Subjects’ positions were logged using a particle filter-based body tracker with head pose estimation \cite{lanz2006approximate}. Conversational groups were annotated at 5 sec. intervals, resulting in 320 frames with ground truth group  annotations.
    \item[--] \textit{SALSA Dataset} \cite{SALSA}. 18 participants were recorded using multiple cameras and sociometric badges and then annotated at 3 second intervals over the course of 60 minutes, giving 1,200 total frames. The dataset consists of a \textit{poster presentation} session and a \textit{cocktail party}. Despite the differences in the structure of F-Formations that appear in these two settings, we treat SALSA as a single dataset to test generalization to different group formations. 
    \item[--] \textit{Coffee Break Dataset} \cite{cristani2011social}. Images were collected using a single camera outdoors. People engaged in small group conversations during coffee breaks. The number of people per frame varied from 6 to 14. People tracking is rough, with orientations only taking values of 0, 1.57, 3.14, and 4.71 radians. Compared to Cocktail Party and SALSA, the spatial features provided by Coffee Break are far noisier. A total of 119 frames have ground truth group annotations.
   
\end{itemize}

\subsubsection{Data Augmentation}

Due to the small size of the datasets, we augment them by rotating position and orientation data by 180 degrees about the world coordinate frame W, giving twice as many training, validation, and test examples. Since groups are define based on person ID’s, they do not need to be adjusted for the augmented data. This augmentation corresponds to tracking the participants from the opposite side of the room/space and is therefore semantically valid. We report all accuracies using the augmented datasets. 

Additionally, we double the size of our augmented datasets by flipping position and orientation data over the vertical axes of W, which allows us to predict $a_{ij}$ and $a_{ji}$, as described in Section 3.3. Note that this augmentation doubles the amount of training data available to DANTE, but does not change the total number of training, validation, or test examples on which we report our evaluation metrics, because we average the pair of predicted affinities for each example.


\subsection{Evaluation Metrics}

We consider standard evaluation metrics for conversational group detection \cite{cristani2011social, setti2013multi, setti2015f, vascon2016detecting,Sanghvi2018-rssw}.
A given group $k$ is said to be correctly estimated if $\ceil*{T*|g_k|}$ of their members are correctly estimated and if no more than $1 - \ceil{T*|g_k|}$ false subjects are identified, where $|g_k|$ is the cardinality of the labeled group $g_k$ and $T$ is a defined tolerance threshold. Common values of $T$ are $2/3$ and $1$ \cite{cristani2011social, setti2013multi, setti2015f, vascon2016detecting, vazquez2015parallel, Sanghvi2018-rssw}. We center our attention on evaluating methods based on $T=1$, i.e., on perfect group detection, since it is more challenging than $T=2/3$.

Let $TP$ (true positive) to be a correctly detected group, $FN$ (false negative) be a non-detected group, and $FP$ (false positive) be a group that was detected but did not exist. Then, we measure our accuracy with three metrics: precision, recall, and $F_1$ score. Precision is defined as $Prec = \frac{TP}{TP + FP}$, recall is $Rec = \frac{TP}{TP + FN}$, and $F_1$ score is $F_1 = 2 \cdot \frac{Prec \cdot Rec}{Prec + Rec}$.

\subsection{Baselines}

We focus on comparing the proposed conversational group detection  approach against two state-of-the-art methods. First, we compare our proposed approach against the graph-cuts method by Setti et al. \cite{setti2015f} (GCFF), given that it outperforms prior model-based group detection approaches, e.g., \cite{cristani2011social, setti2013multi}. Second, we compare results against the game theoretic approach of Vascon et al. \cite{vascon2016detecting} (GTCG)  because this method relies on graph clustering, like our approach. Also, \cite{vascon2016detecting} tends to give better performance than the alternative graph-based approach by Hung and Kr{\"o}se \cite{hung2011detecting}. 

Although we did not re-run the group communication approach by Sanghvi, Yonetani, and Kitani \cite{Sanghvi2018-rssw} (GComm), we report results from their publication as a reference. This approach is of interest as well because it uses deep learning for conversational group detection. Note that we omit comparisons against \cite{hedayati2019recognizing}, which also uses machine learning, because in our experimental setup this approach resulted in worse performance than the other state-of-the-art baselines with the $T=1$ F1  evaluation criteria.

\subsection{Experimental Setup}

Due to the small size of the datasets, we use 5-fold cross validation to measure performance and study variability in the results. Each fold is taken as a continuous section of data due to the inherent auto-correlation of time-series spatial features from the datasets' videos. We select data for validation from the training set such that it separates as much as possible the data that is actually used for training from the one that is used for testing. The test data of a given fold is only used for computing final results after hyper-parameters are chosen based on the validation set.

In order to fairly compare our results against previous work, we fine-tuned the state-of-the-art baselines \cite{setti2015f, vascon2016detecting} using the corresponding training and validation data for each fold. The average results for the graph-cuts approach of Setti  et al. were slightly improved in comparison to \cite{setti2015f}. Note that \cite{vascon2016detecting} does not present results for $T=1$ F1.

To train \methodname, we use the log loss between each predicted affinity and the true $\{0, 1\}$ affinity, corresponding to whether or not two people are part of the same conversational group. We optimize the network through gradient descent with the Adam optimizer \cite{Kingma2014AdamAM}, a learning rate of 0.0001 and a batch size of 64 samples. We search for the best  hyper-parameters using the validation data, including the number of layers in \methodname's multi-layer perceptrons ($D_{\textit{dyad}}, D_{\textit{context}}, D_{\textit{comb}}$) as well as the size of these layers. 

\begin{table*}[b]
\centering
\caption{Results on various conversational group detection benchmarks. The scores are $T=1$ F1 values averaged across five folds. See the Appendix for a break-down of the results by each fold.}
\label{tab:f1-results}
\begin{tabular}{lccccccc}
\cline{1-7}
Method & Cocktail Party  & Coffee Break  & SALSA &  \\ \cline{1-7}
GComm \cite{Sanghvi2018-rssw} & 0.60 & 0.63 & -  \\
GTCG \cite{vascon2016detecting} &  0.29 & 0.48 & 0.44  \\
GCFF \cite{setti2015f} & 0.64 & 0.63 & 0.41 \\
DANTE & \textbf{0.73} & 0.64 & \textbf{0.65} \\
DANTE-NoContext & 0.64 & \textbf{0.66} & 0.57 \\ 
\cline{1-7}
\end{tabular}
\end{table*}


\subsection{Quantitative Results -- Comparison Against Baselines}

Table \ref{tab:f1-results} shows quantitative results for the Cocktail Party, SALSA, and Coffee Break datasets. The results correspond to $T=1$ F1 scores, averaged across all five folds.  
The Appendix provides  expanded results per fold.

On average, our proposed approach outperforms the F1 scores of the baselines in the Cocktail Party dataset (see DANTE row in Table \ref{tab:f1-results}). The average improvement is 9\% over GCFF, 13\% over GComm, and 44\% over GTCG. We conducted pairwise t-tests comparing per fold results of our approach against GTCG, and found that DANTE led to significantly higher F1 scores ($t(6.88)=5.08$, $p=0.002$). The difference between DANTE and GCFF was not statistically different ($p=0.28$) but DANTE clearly outperformed  GCFF in the first and second folds (see Table  \ref{tab:results-cparty} in the Appendix). These results  suggest that our approach improves the state of the art in conversational group detection when the input spatial data has reasonable quality.

Our proposed approach significantly outperforms GTCG ($t(5.36)=3.93$, $p=0.01$) and GCFF ($t(6.7)=2.59$, $p=0.04$) in the SALSA dataset; the comparison was not possible against GComm because \cite{Sanghvi2018-rssw} does not evaluate on the SALSA dataset. Across all folds, our method results in higher F-1 scores than the baselines (see Table \ref{tab:results-SALSA} in the Appendix). This finding suggests that our data-driven approach can handle the different types of group formations observed in SALSA.

The benefits of our approach are less noticeable in the Coffee Break dataset, where our method performs as well as the best baselines (GCFF \& GComm). Although DANTE leads on all folds in terms of F1 score, it underperforms in Fold 1, as can be seen in the Appendix (Table \ref{tab:results-cbreak}). One reason for this discrepancy is that Fold 1 had the most noisy spatial features $\mathbf{f}$ for the individuals in the scene. This  hurt prior work, but was especially harmful to our data-driven method.

\newcolumntype{C}[1]{>{\centering\let\newline\\\arraybackslash\hspace{0pt}}m{#1}}
\newcommand*{\mysizea}{0.265}
\newcommand*{\mysizeb}{0.20}
\begin{figure*}[t!p]
\centering
\begin{tabular}{ C{3.4cm}C{3cm}C{3cm}C{3cm} } 
 Scene & Ground Truth & GCFF & DANTE
\end{tabular}
\begin{subfigure}[b]{\linewidth}%
\raisebox{0mm}{\includegraphics[frame,width=\mysizea\linewidth]{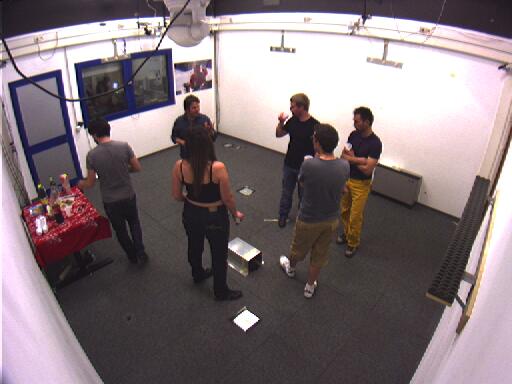}}%
\hspace{0.8em}%
\raisebox{1.2mm}{\fbox{\includegraphics[trim={0.4cm 1cm 0 .2cm}, clip,width=\mysizeb\linewidth]{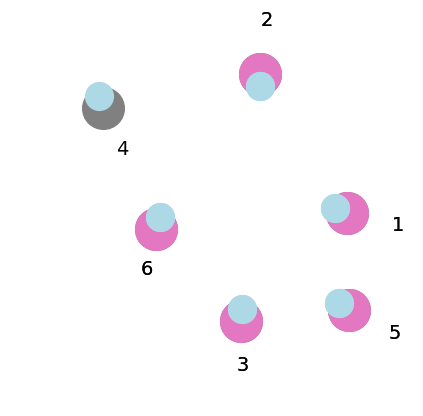}}}\hspace{0.8em}%
\raisebox{1.2mm}{\fbox{\includegraphics[trim={0.4cm 1cm 0 .2cm}, clip,width=\mysizeb\linewidth]{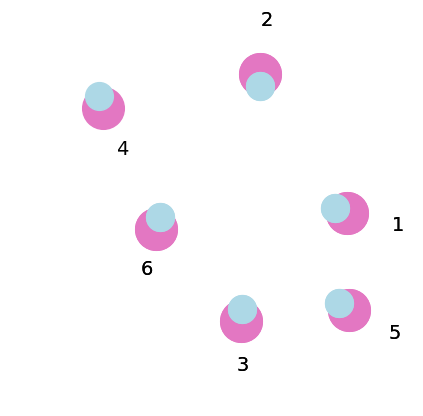}}}\hspace{0.8em}
\raisebox{1.2mm}{\fbox{\includegraphics[trim={-0.0cm .6cm -.3cm .2cm},clip,width=\mysizeb\linewidth]{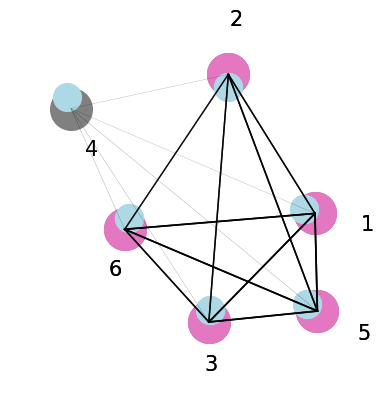}}}%
\end{subfigure}%
\\
\begin{subfigure}[]{\linewidth}%
\raisebox{0mm}{\includegraphics[frame,width=\mysizea\linewidth]{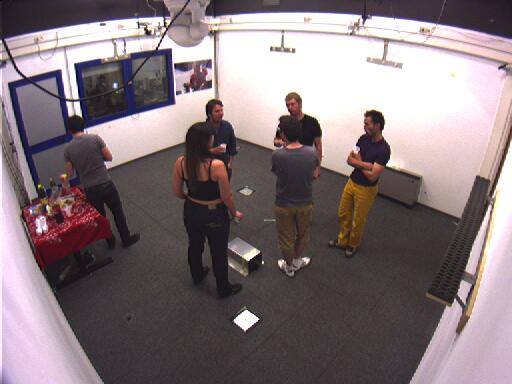}}%
\hspace{0.8em}%
\raisebox{1.1mm}{\fbox{\includegraphics[trim={0.5cm 1.4cm 0 .7cm},clip,width=\mysizeb\linewidth]{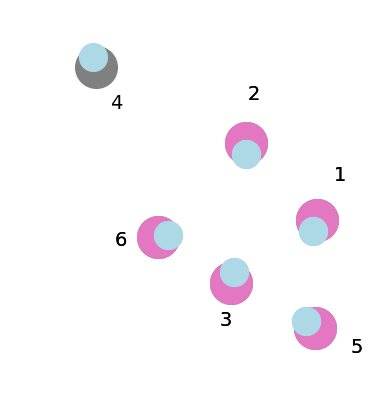}}}\hspace{0.8em}%
\raisebox{1.1mm}{\fbox{\includegraphics[trim={0.5cm 1.4cm 0 .7cm}, clip,width=\mysizeb\linewidth]{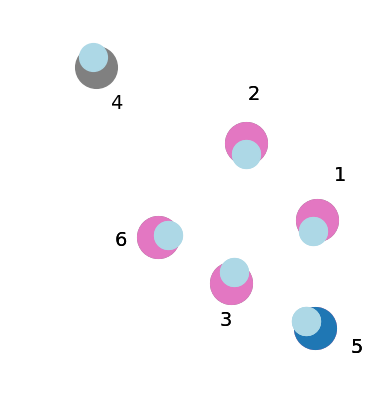}}}\hspace{0.8em}%
\raisebox{1.1mm}{\fbox{\includegraphics[trim={-0.4cm .8cm 0.0cm 0.6cm},clip,width=\mysizeb\linewidth]{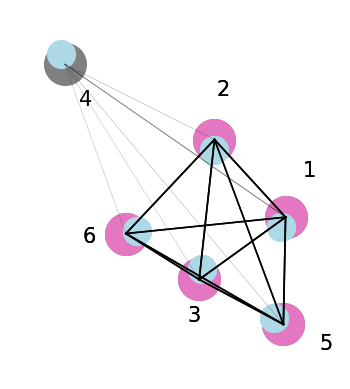}}}%
\end{subfigure}%
\\%
\begin{subfigure}[h]{\linewidth}%
\raisebox{1.2mm}{\includegraphics[frame,width=\mysizea\linewidth]{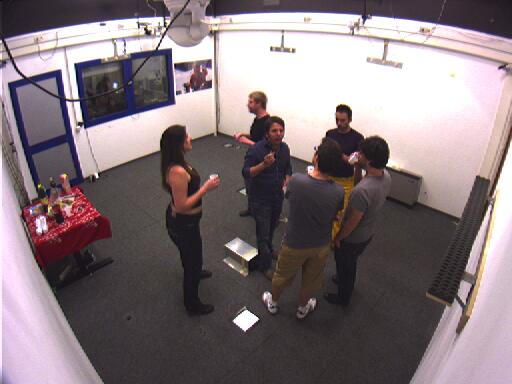}}%
\hspace{0.8em}%
\raisebox{2.4mm}{\fbox{\includegraphics[trim={1.5cm 1.2cm 0cm 1.1cm}, clip,width=\mysizeb\linewidth]{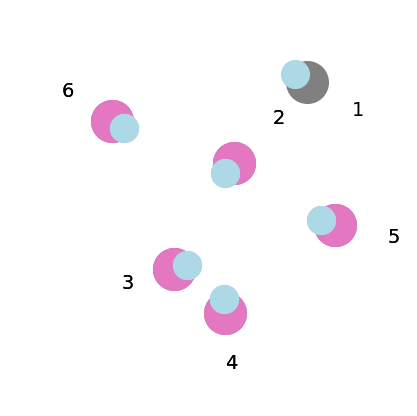}}}\hspace{0.8em}%
\raisebox{2.4mm}{\fbox{\includegraphics[trim={1.5cm 1.2cm 0cm 1.1cm}, clip,width=\mysizeb\linewidth]{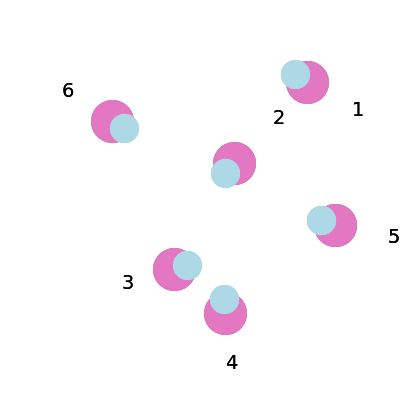}}}\hspace{0.8em}%
\raisebox{2.4mm}{\fbox{\includegraphics[trim={.35cm 0.3cm 0cm 1.1cm},clip,width=\mysizeb\linewidth]{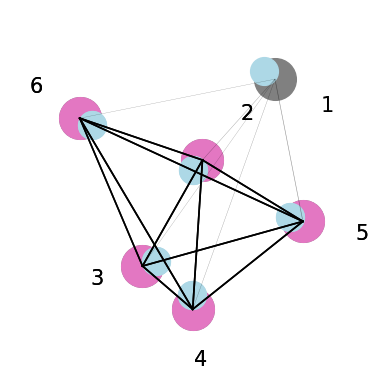}}}%
\end{subfigure}%
\\ \vspace{-0.3em}%
\begin{subfigure}[h]{\linewidth}%
\raisebox{0mm}{\includegraphics[frame,width=\mysizea\linewidth]{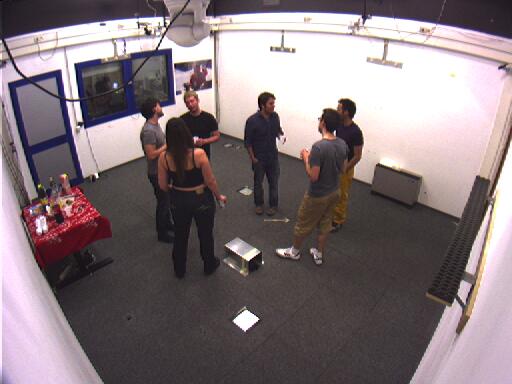}}%
\hspace{0.8em}%
\raisebox{1.3mm}{\fbox{\includegraphics[trim={.3cm 1.1cm 0cm .2cm},clip,width=\mysizeb\linewidth]{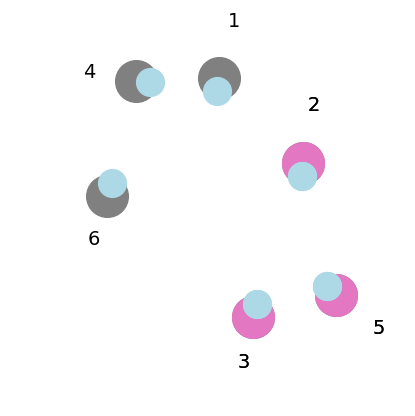}}}\hspace{0.8em}%
\raisebox{1.3mm}{\fbox{\includegraphics[trim={.3cm 1.1cm 0cm .2cm},clip,width=\mysizeb\linewidth]{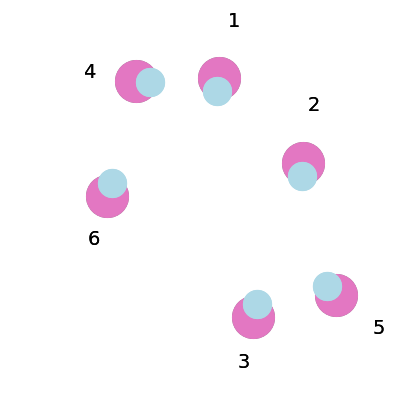}}}\hspace{0.8em}%
\raisebox{1.3mm}{\fbox{\includegraphics[trim={-.43cm .6cm 0cm .2cm},clip,width=\mysizeb\linewidth]{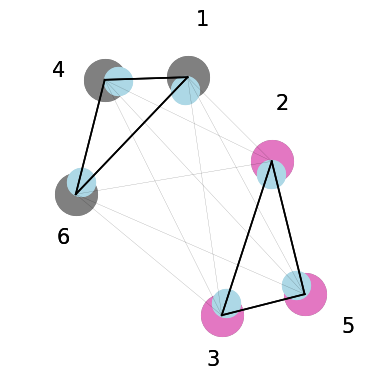}}}%
\end{subfigure}%

\caption{Example results. \textit{First Column:} Original image from the Cocktail Party dataset, \textit{Second Column:} ground truth conversational group, \textit{Third Column:} results from GCFF \cite{setti2015f}, \textit{Fourth Column:} our results with DANTE. The wall with the door in the images corresponds to the \textbf{top} side of the diagrams in the second to fourth columns. People's colors indicate groups, and line thickness indicates DANTE's affinity prediction (thicker means closer to 1 in the [0,1] range). In comparison to DANTE, GCFF tends to be more inclusive, resulting in incorrect large groups. This result aligns  with prior findings \cite{vazquez2015parallel}.}
\vspace{-1em}
\label{fig:analysis}
\end{figure*}

\subsection{Qualitative Results}

\begin{figure}[t!p]
\centering
\begin{subfigure}[h]{.85\linewidth}%
\includegraphics[frame,width=0.381\linewidth]{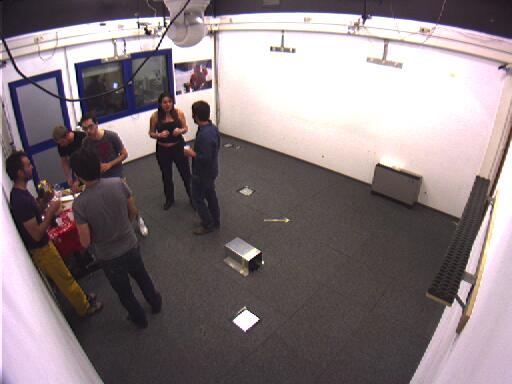}%
\hspace{0.4em}%
\raisebox{1.2mm}{\fbox{\includegraphics[trim={.2cm .6cm .2cm 0.22cm},clip,width=0.27\linewidth]{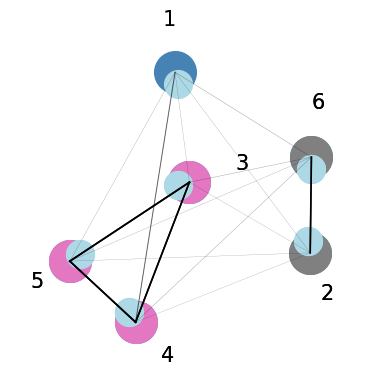}}}\hspace{0.4em}%
\raisebox{1.2mm}{\fbox{\includegraphics[trim={.2cm .6cm .2cm 0.22cm},clip,width=0.27\linewidth]{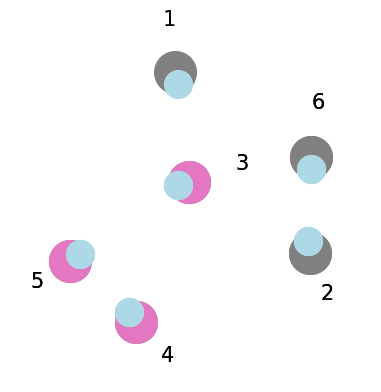}}}%
\caption{DANTE estimates that the person 1 does not belong to a conversational group while he grabs an object from the coffee table. Ground truth groups: $g_1=\{1, 3, 4, 5\}, g_2=\{2, 6\}$. }%
\vspace{10pt}
\end{subfigure}%
\\
\begin{subfigure}[h]{.85\linewidth}%
\includegraphics[frame,width=0.381\linewidth]{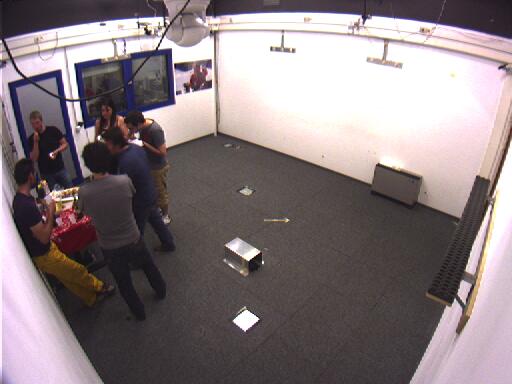}%
\hspace{0.4em}%
\raisebox{1.2mm}{\fbox{\includegraphics[trim={-1cm 0 -1cm .4cm},clip,width=0.27\linewidth]{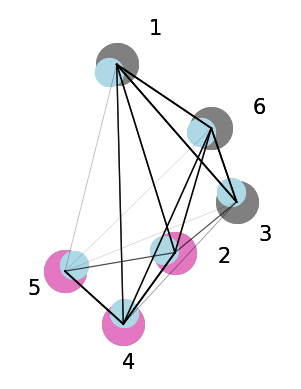}}}\hspace{0.4em}%
\raisebox{1.2mm}{\fbox{\includegraphics[trim={-1cm 0 -1cm .4cm}, clip,width=0.27\linewidth]{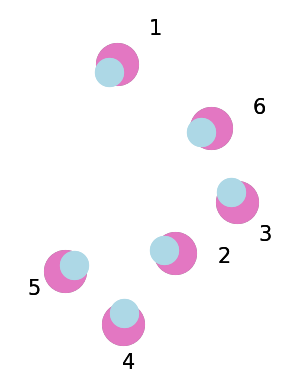}}}%
\caption{DANTE estimates two groups while all individuals are conversing together. Ground truth groups: $g_1=\{1, 2, 3, 4, 5, 6\}$.}%
\end{subfigure}%
\caption{Failure cases by the proposed  approach. \textit{Left:} original image from the Cocktail Party dataset, \textit{Middle:} estimated groups by DANTE, \textit{Right:} ground truth. Individuals are colored based on their groups in each case.}  
\label{fig:failures}
\vspace{-0.5em}
\end{figure}

 Figures \ref{fig:analysis} and \ref{fig:failures} provide qualitative results in the Cocktail Party dataset. In particular,  Fig.~\ref{fig:analysis} compares example results between our method and the GCFF approach \cite{setti2015f}. In rows 1, 3, and 4, GCFF chooses larger groups due to a penalty on small group sizes. This preference for larger groups often overrides information in the data, such as a person facing away from the proposed group. In row 2, Person 5 is likely excluded from GCFF's primary group due to a heuristic in the GCFF algorithm which prevents grouping two people if there is someone else in-between them. In comparison, our deep learning approach considers the social context of the groups to learn more nuanced spatial patterns, such as how one orients oneself when leaving a group (row 1), how one behaves when standing on the outskirts (rows 2 and 3), and how people arranged in a ring can still form smaller groups (row 4). This flexibility largely comes from not employing brittle heuristics to account for context and instead allowing the model to learn from data. Also note that the edge-weights are either very correctly thick due to high predicted affinities (> .9) or correctly thin (< .2) in these examples. In general, we found DANTE to be very confident when making successful predictions.

Figure~\ref{fig:failures} shows failure cases by our method. In Fig. \ref{fig:failures}(a), one of the main limitations of our approach becomes evident: useful information (e.g. posture or gaze) to correctly assign the individual 1 to the right group is not available to DANTE using the default spatial features described in Sec. \ref{sec:implementation}. Our method only has access to 2D position and orientation in the Cocktail Party dataset, which can make some interaction analysis difficult.
In case (b), another limitation is apparent: DANTE lacks environmental features (e.g. table or wall locations), which could explain the large space in between the two predicted groups.  Instead, DANTE infers this large empty space to signify that the two cohorts are separate conversations. Particularly in case (b), DANTE predicts many intermediate affinity values with inconsistent weights across some node clusters, an indication of the uncertainty in the affinity predictions and uncertainty in determining the group structures. These failure cases illustrate opportunities for future improvement, as further discussed in Section \ref{sec:discussion}.

\begin{figure}[t!p]
\centering
\begin{subfigure}[c]{.85\linewidth}%
\includegraphics[frame,width=0.381\linewidth]{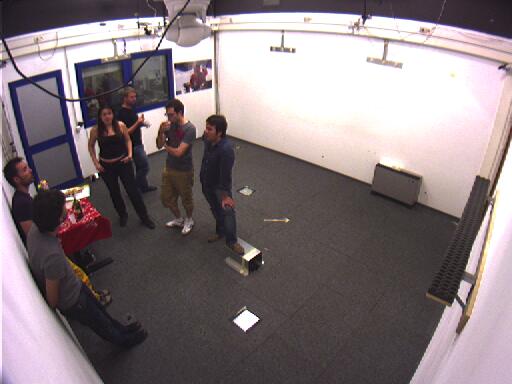}%
\hspace{0.4em}%
\raisebox{1.2mm}{\fbox{\includegraphics[trim={0cm 1cm 0cm .5cm},clip,width=0.27\linewidth]{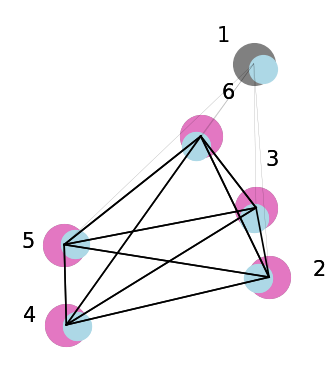}}}%
\hspace{0.4em}%
\raisebox{1.2mm}{\fbox{\includegraphics[trim={0cm 1cm 0cm .5cm},clip,width=0.27\linewidth]{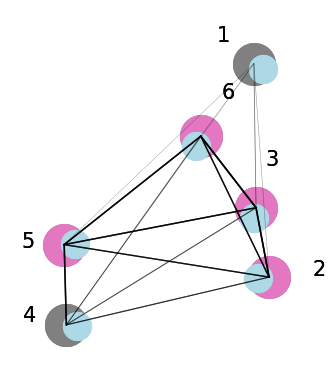}}}%
\caption{Without global context, DANTE-NoContext groups people $1$ and $4$, even though they are clearly occluded. Ground truth groups: $g_1=\{2, 3, 4, 5, 6\}, g_2=\{1\}$ (not interacting).}
\vspace{10pt}
\end{subfigure}%
\\
\begin{subfigure}[c]{.85\linewidth}%
\includegraphics[frame,width=0.381\linewidth]{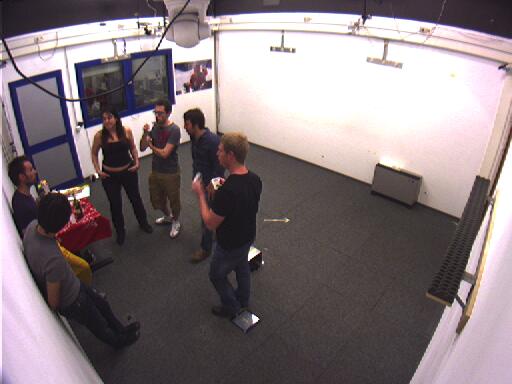}%
\hspace{0.4em}%
\raisebox{1.2mm}{\fbox{\includegraphics[trim={.4cm .4cm .4cm .4cm},width=0.271\linewidth]{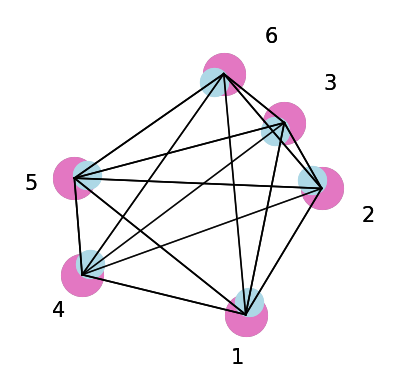}}}%
\hspace{0.4em}%
\raisebox{1.2mm}{\fbox{\includegraphics[trim={.4cm .4cm .4cm .4cm},width=0.271\linewidth]{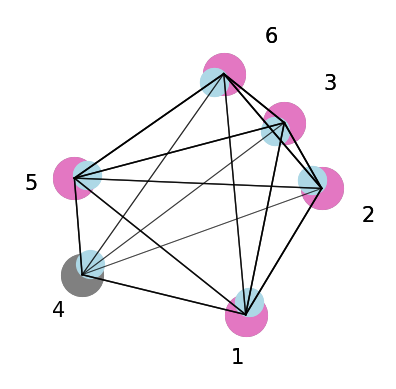}}}%
\caption{DANTE accounts for the large group when computing pairwise-affinities, while DANTE-NoContext gives Person 4 lower pairwise-affinities based on distance and an inability to notice the large group. Ground truth groups: $g_1=\{1, 2, 3, 4, 5, 6\}$. }
\end{subfigure}%
\caption{Ablative analysis.
\textit{Left:} Image from the Cocktail Party dataset, \textit{Middle:} DANTE, \textit{Right:} DANTE-NoContext. Grouped individuals share the same color.}
\label{fig:cases1}
\vspace{-0.6em}
\end{figure}

\subsection{Analysis of the Effect of the Context Transform in DANTE}

We hypothesized that adding contextual information to the affinity computation by DANTE would improve the results of our group detection approach. To explore if this was effectively the case, we performed a small ablation study. In particular, we evaluated a version of DANTE that only reasoned about the position and orientation of the individuals of interest using the Dyad Transform. 

The results for the Cocktail Party dataset are presented in Table \ref{tab:f1-results} in the row corresponding to DANTE-NoContext. As expected, excluding the Context Transform from DANTE resulted in 9\% worse average group detection performance, although a t-test on the difference in F1 scores between DANTE and DANTE-NoContext was not significant (p=0.39). We attribute the lack of significance to results being similar in two out of the five folds (see Table \ref{tab:results-cparty} in the Appendix). 

Figure \ref{fig:cases1} shows example, qualitative results in the Cocktail Party dataset. Comparing DANTE vs. DANTE-NoContext, we can observe that the social context input to the affinity computation is highly relevant to the group detection task. In Fig.~\ref{fig:cases1}(a), two people that are separated by another interaction are grouped with one another, even though this would be unlikely in real situations. In real life, the two people would have trouble communicating with each other when a conversation is happening in-between them. Although it may seem surprising that persons 1 and 4 could be given high affinity, there were large groups in the datasets with members at similar distances. The context of the other people in the room thus became necessary to differentiate these cases. In Fig.~\ref{fig:cases1}(b), a person is missed in a group interaction, due to slightly lower affinities than the rest of the pairwise interactions. Overall, we see that without context, DANTE is often unsure of what to do with people at greater distances, and these less confident affinities can lead to incorrect groupings. This suggests that our complete version of DANTE is better at reasoning about complex spatial patterns.  

The results in the SALSA dataset were similar to the Cocktail Party. We found that removing the Context Transform from DANTE led to reduced average F1 scores across all folds (0.65 for DANTE vs. 0.57 for DANTE-NoContext). Although DANTE was consistently better with the context information, a t-test resulted in no significant differences  (p=0.33). 

In Coffee Break, the average group detection results were slightly better without the Context Transform (Table 1). Although the difference was not significant (p=0.86), the results surprised us. We attribute the lack of benefit of the context in this case to the more noisy annotations and less data provided by this dataset. This idea is supported by the fact that DANTE performs comparatively worse than DANTE-NoContext on the noisiest fold (Fold 1 in Table \ref{tab:results-cbreak} within the Appendix). We believe that this is due to the increased complexity of DANTE vs DANTE-NoContext, which causes it to overfit on the training data. Worth noting, DANTE-NoContext slightly outperformed other baselines in terms of average F1 score in this benchmark. This finding reinforces the idea that deep learning can help with the group detection task.

\section{Generalization Experiment}
\label{sec:generalization_experiment}
Although our focus is on conversational group detection, we performed an experiment using a large, general group detection dataset. While  prior conversational group detection algorithms often relied on heuristics tailored to conversational group detection \cite{yu2009monitoring, hung2011detecting, vascon2016detecting, setti2015f, vazquez2015parallel}, we hoped that our data-dependent procedure would allow it to perform well outside of its initially intended domain. We followed the same data-augmentation and setup procedures as in Sec. \ref{sec:experiments} for this experiment.



\subsection{Dataset}
\textit{Friends Meet} \cite{FriendsMeet} is composed of 53 synthetic and real sequences of varying group types, including but not restricted to conversational groups. Keeping in line with prior work \cite{VASCON2017}, we restrict our training and evaluation to the synthetic sequences. These sequences were chosen by \cite{VASCON2017} because the real sequences are not labeled by group type. Also, \cite{VASCON2017} removed queuing sequences from the data because queues are semantically and spatially different from the other group interactions in the dataset, e.g., groups of pedestrians that walk together towards a destination. Therefore, we present our results based on the 25 non-queuing synthetic sequences, with 200 annotated frames per sequence, for a total of 5,000 frames. 

Because Friends Meet only provides position information for potential interactants, we perform the same feature augmentation as \cite{VASCON2017}, which subtracts people's position between consecutive frames to obtain crude velocity estimates. These estimates are normalized and used in place of the orientation features considered in \methodname's default input representation (Sec. \ref{sec:implementation}). Using the  
velocity instead of the orientation can be thought of as assuming that people orient in their motion direction. Although this assumption breaks down when individuals are standing still.

\subsection{Evaluation Metrics}
We consider both the challenging $T=1$ F1 metric, as well as the more lenient Group Detection Success Rate (GDSR) employed in prior work \cite{VASCON2017}. GDSR measures the percentage of groups correctly identified, and a group is considered correct if at least 60\% of its members are detected.

\subsection{Results}

The results are presented in Table \ref{tab:results_overall_FM}. Our approach (\methodname\ row) achieved state-of-the-art results on this general group detection benchmark \cite{FriendsMeet}. As in Sec. \ref{sec:experiments}, we conducted independent t-tests to compare methods in terms of their F1 and GDSR scores across folds. In comparison to GTGC \cite{vascon2016detecting}, our approach led to significantly higher F1 scores (t(6.05)=-9.79, p<0.001) as well as higher GDSR, (t(4.12)=-5.71, p=0.004). There was also a significant difference between the F1 scores of GCFF \cite{setti2015f} and our approach, (t(4.82)=2.91, p=0.03). Lastly, the difference in terms of GDSR for the latter two methods was close to significant (p=0.08).



\begin{table}[t!p]
    \caption{Generalization Results on Friends Meet Dataset}
    \centering
    \begin{tabular}{ccc}
    \cline{1-3}
    \label{tab:results-FM}
Method & $T = 1$ & GDSR  \\ \cline{1-3}
GTCG \cite{vascon2016detecting} & 0.66 & 0.90  \\
GCFF \cite{setti2015f} & 0.79 & 0.95 \\
DANTE & \textbf{0.90} & \textbf{0.99} \\
\cline{1-3}
    \end{tabular}
    \label{tab:results_overall_FM}
    \vspace{-0.5em}
\end{table}

\section{Applications}
\label{sec:applications}

Our group detection approach can be used to increase the social awareness of interactive systems. To demonstrate this in practice, we built an interactive system using the Robot Operating System (ROS) \cite{quigley2009ros}, a popular collection of tools and libraries for robotics development.\footnote{\url{https://sites.google.com/view/dante-group-detection}
} This effort adds to a larger body of work that shows that automatic group detection is relevant in HCI, ranging from mobile systems \cite{marquardt2012cross} to robotic interfaces \cite{ichino2016effects, dim2015automatic}.

\subsection{Group Detection for Human-Robot Interaction}
\begin{figure}[t!p]
\includegraphics[width=\linewidth]{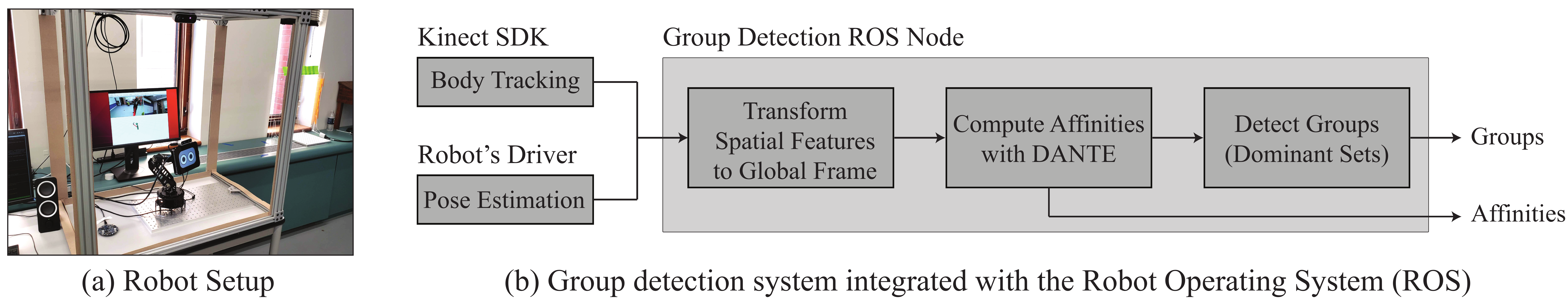}
\caption{\textit{Left:} Robot setup. \textit{Right:} Components of our interactive system.}
\label{fig:ros_system}
\end{figure}

In our demonstration application, a table-top robot is used to identify F-Formations based on users' spatial behavior relative to each other and its own spatial configuration in our lab environment (Figure \ref{fig:ros_system}).  The main components of our interactive system are a robot arm with a screen face, and two RGB-D cameras (Fig. \ref{fig:ros_system}a). The robot and all sensors are connected to a nearby desktop computer, which processes data in real-time and controls the robot. The computer has an NVidia GeForce GTX 1080 Ti graphics card that we use for testing \methodname. 

\subsubsection{Scenario}
The robot, Shutter, acts as a photographer in our demonstration. The robot uses a forward facing RGB-D camera to detect faces such that when an individual or a group approaches the robot, it turns towards them and asks them if they would like to be photographed. If they accept, Shutter then counts down from 3 and takes a photo. 
While Shutter interacts with people, we use the secondary RGB-D camera that is fixed above the robot to reason about groups. This camera is a Kinect Azure. It provides a wider view of the environment, augmenting the robot's own sensors.

\subsubsection{Spatial Features}
As illustrated in Fig. \ref{fig:ros_system}b, we use the fixed camera above the robot for human body tracking in real-time via the Azure Kinect Body Tracking SDK \cite{kinect_azure}. The Kinect SDK provides keypoint tracking information for each person in view. This information includes position and head orientation, which we use for spatial reasoning. 

We consider the robot as a social agent in the demonstration scenario because it socially engages in conversations with users. For the robot's spatial features, we use its position on the table, which is known a priori and fixed based on our setup. For its head orientation, we use the orientation of its screen face, which is obtained from its joint positions.  

\subsubsection{Real-Time Group Detection}
All the group detection logic is encapsulated into a ROS node, a computer process that runs within the ROS network of the robot (Figure \ref{fig:ros_system}b). At a given time, the group detection node processes spatial information as follows. First, it transforms the latest vision-based person tracking information and the robot's pose information that has been received into a shared, global coordinate frame. The positions and orientations are then organized into an interaction graph, so that affinities can be computed for all pairs of social agents using \methodname\ (pre-trained on the Cocktail Party dataset used for the evaluation of Sec. \ref{sec:experiments}). Then, the affinities are processed by the Dominant Sets algorithm, which finally outputs groups. Note that our system also outputs affinity scores for visualization purposes, as illustrated in Figures  \ref{fig:shutter_excluded} and \ref{fig:separate_group}. In these Figures, the affinity values are displayed as red lines between the social agents in the scene. The opacity of the lines indicates the strength of the affinities (more opaque means stronger).
In the middle and right images of Figures  \ref{fig:shutter_excluded} and \ref{fig:separate_group}, each detected person is denoted with a cylindrical marker, while Shutter is depicted as a 3D model of the platform. The people markers are colored according to the group to which they belong. That is, people predicted to be within the same group are of the same color. Shutter also has a colored arrow above it that indicates its group affiliation.

\subsubsection{Qualitative Results}
\begin{figure}[t!p]
\minipage{0.32\textwidth}
  \includegraphics[width=\linewidth]{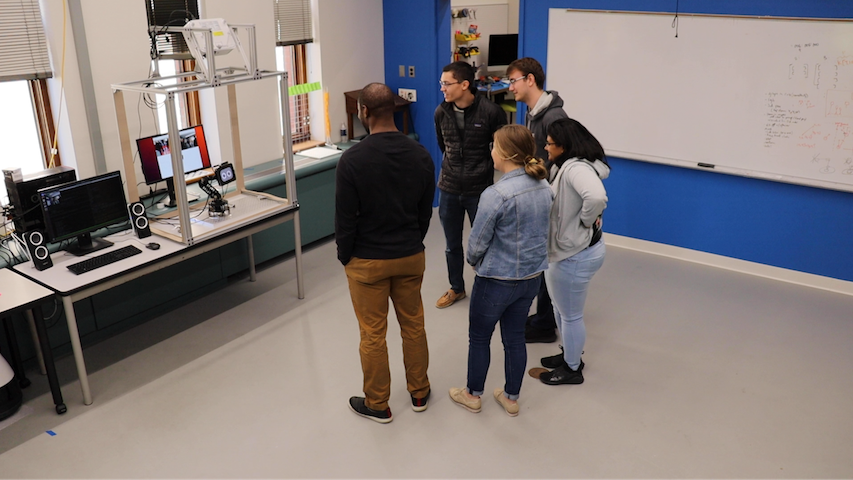}
\endminipage\hfill
\minipage{0.32\textwidth}
  \includegraphics[width=\linewidth]{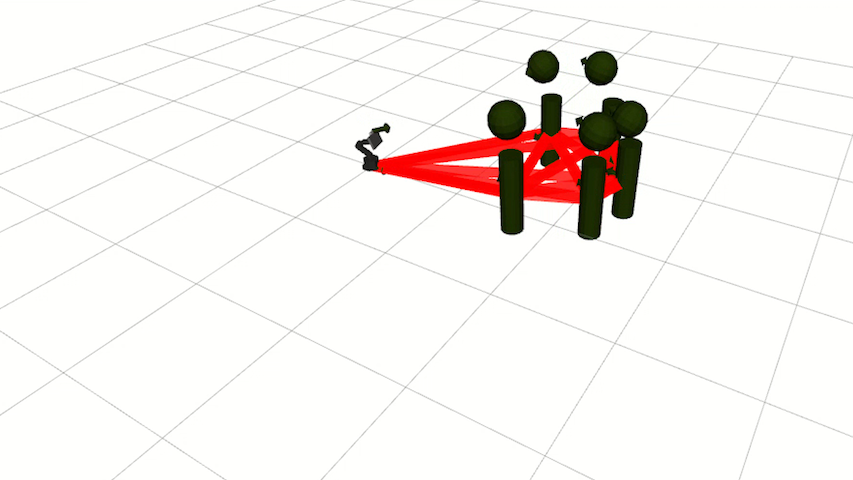}
\endminipage\hfill
\minipage{0.32\textwidth}%
  \includegraphics[width=\linewidth]{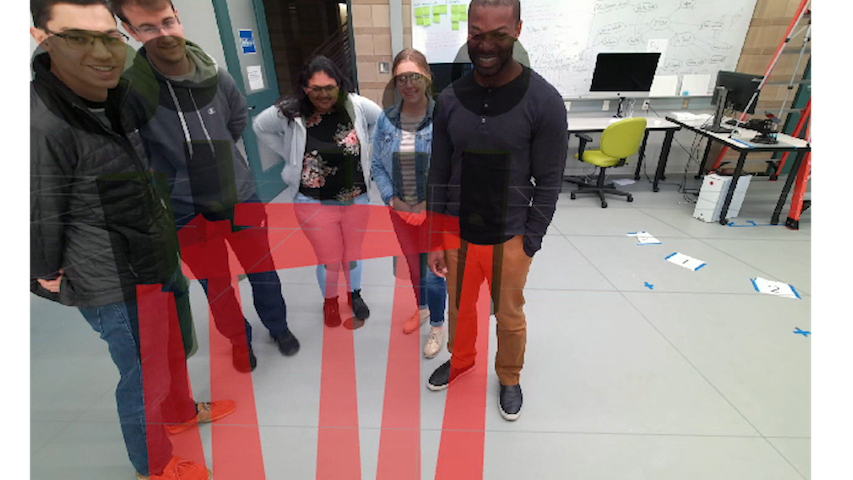}
\endminipage
\caption{Shutter interacting with a group after taking a photo and talking  to users. All markers are colored dark green in the middle image, indicating that all agents belong to one group. Best viewed in digital form.}
\label{fig:shutter_interaction}
\end{figure}

We investigate the performance of DANTE under three scenarios: groups including Shutter, groups excluding Shutter, and multiple groups within the same frame.  Figure \ref{fig:shutter_interaction} shows an example of the first case, in which Shutter interacts with a group of five individuals. The pairwise affinities between each social agent are strong, corresponding to opaque lines connecting the agents. Overall, our method worked well in these types of scenarios, except when Shutter moved its face to one side and suddenly the group divided in two. Sometimes the split was correct. The robot was trying to interact with a single individual, whose face was detected by its camera. But sometimes the motion was due to the robot turning around during part of the interaction to signal to its users that they should look at the screen behind it to see their picture. Our group detection system had no information about this change in activity, but this information could be considered by it in the future.

\begin{figure}[t!p]
\minipage{0.32\textwidth}
  \includegraphics[width=\linewidth]{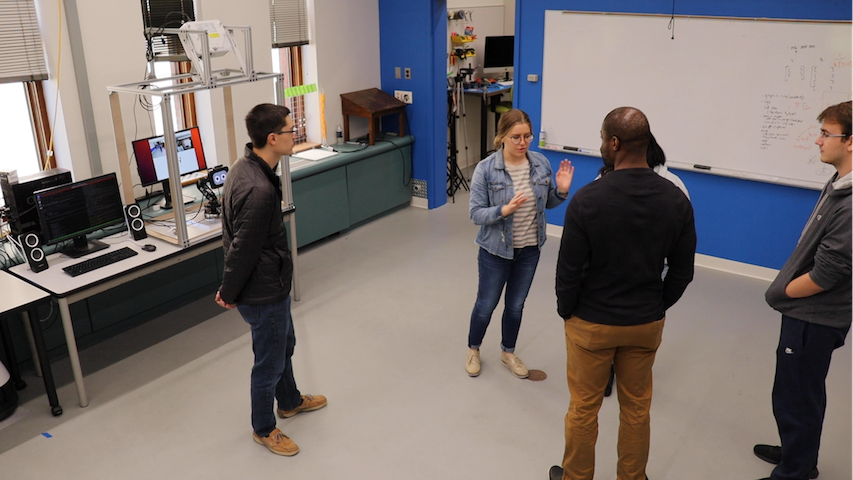}
\endminipage\hfill
\minipage{0.32\textwidth}
  \includegraphics[width=\linewidth]{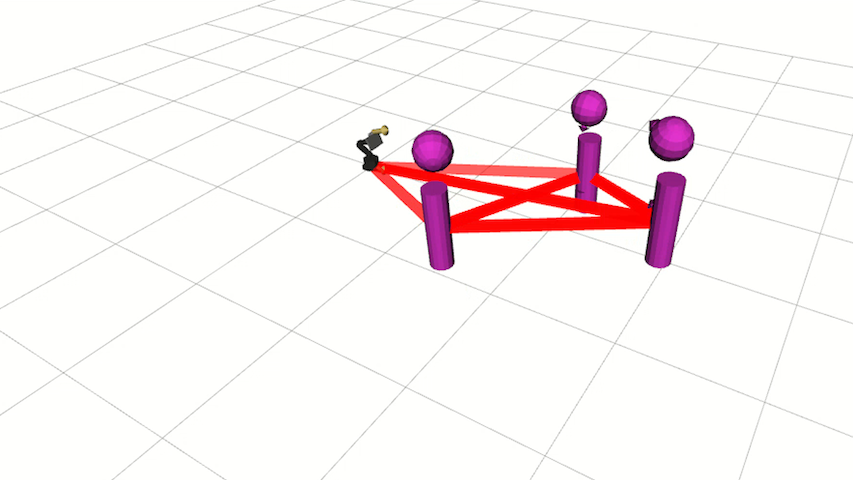}
\endminipage\hfill
\minipage{0.32\textwidth}%
  \includegraphics[width=\linewidth]{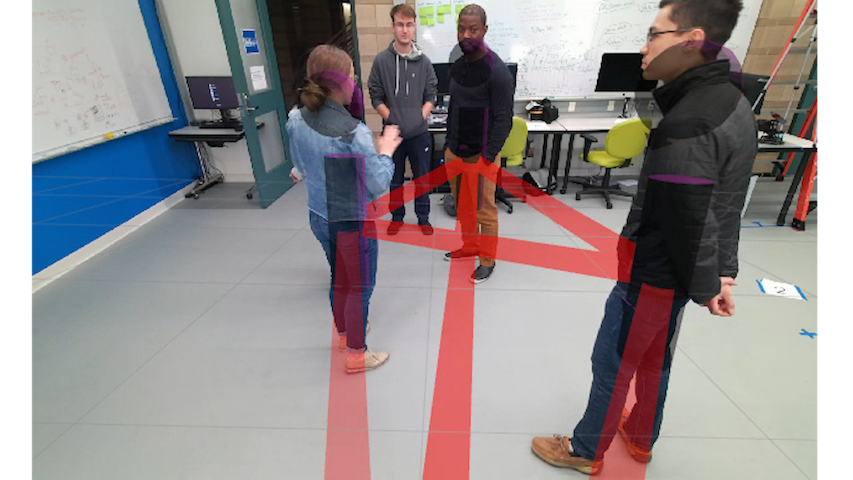}
\endminipage
\caption{Example in which Shutter is excluded from a human group (the robot's marker is yellow in the middle image, while the people are purple). Best viewed in digital form.}
\label{fig:shutter_excluded}
\end{figure}

After  Shutter took a photo, the group moved back and formed a group distinct from Shutter (Figure \ref{fig:shutter_excluded}). In this scenario, Shutter is detected to be in a separate group from the other people in the room. But because of the setup of the Kinect camera, the other two people in the group were occluded and their poses were not detected. Thus, they were not grouped together in this case.

\begin{figure}[t!p]
\minipage{0.32\textwidth}
  \includegraphics[width=\linewidth,trim={0cm 3cm 0 2cm},clip]{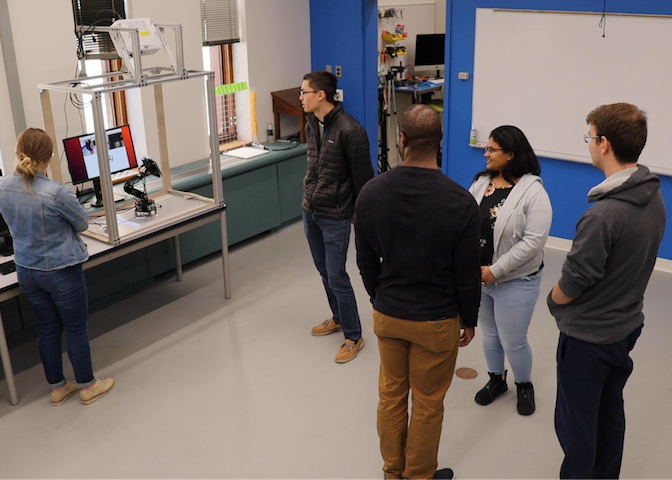}
\endminipage\hfill
\minipage{0.32\textwidth}
  \includegraphics[width=\linewidth]{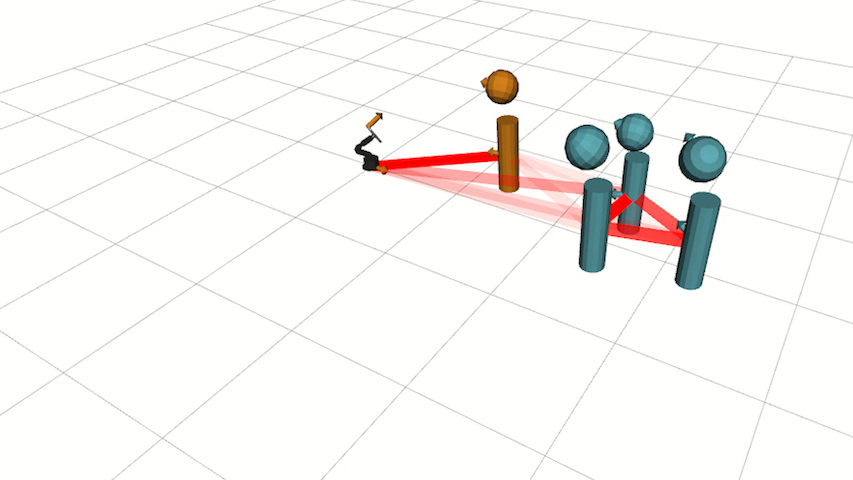}
\endminipage\hfill
\minipage{0.32\textwidth}%
  \includegraphics[width=\linewidth]{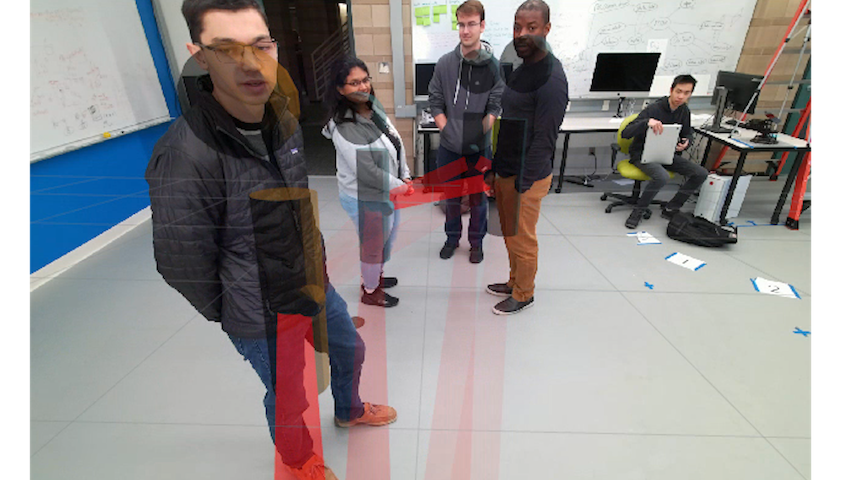}
\endminipage
\caption{Example of Shutter interacting with a single individual while a separate group talks in the background. Shutter's marker is orange - the same color as the person closest to the camera. Best viewed in digital form.}
\label{fig:separate_group}
\end{figure}

In general, the algorithm detected groups distinct from Shutter's, but occasionally had problems in separating the social agents standing in close proximity to the robot. This sometimes happened because the robot was looking toward another group, the  members of a nearby group looked at the robot over their shoulders, or they left space for Shutter to look into their group. 
Considering additional spatial features, such as body and shoulder orientation, during group detection could help reduce the number of errors that occur in these situations.

Our approach is also capable of detecting multiple groups in real-time (often running at 15 Hz, the framerate of the Kinect camera). For example, consider Figure \ref{fig:separate_group}. In this case, one person interacts with Shutter and there is another group in the background. Note that the affinities between the group in the foreground and the background are nonzero, but significantly weaker than the affinities between group members. 
When DANTE misclassified the groups, it was often because the members of one group were looking toward the other, as described before. But overall, the algorithm performed well at separating distinct groups that occurred simultaneously.


\subsection{Other Applications}
We envision using our group detection approach in other applications that benefit from social intelligence. For example, our approach could be used to improve social robot navigation, including delivery and guide robots \cite{triebel2016spencer,evers2014development}, autonomous cars \cite{schneemann2016context}, and assistive wheelchairs \cite{gomez2019affordable}. Additionally, our method could be used to create better embodied conversational assistants in office environments \cite{bohus2009dialog, bohus2017study} or educational settings \cite{matsuyama2016socially}.  Finally, we look forward to testing our approach for creating interactive public installations, like facades that respond to the group activities of nearby pedestrians \cite{dalsgaard2010designing}. All these types of interfaces require increased levels of social intelligence, which we believe our approach can contribute to. Naturally, the performance of our method will be subject to the availability of appropriate training data, even if it is borrowed from another related domain as in our example demonstration.

\section{Discussion}
\label{sec:discussion}

\subsection{Evaluation}
In comparing our method's performance against prior work across datasets, we observed that our model tended to perform better on larger, better annotated datasets. This is promising because reliable spatial features are quickly becoming a commodity, e.g., as provided by the Kinect Azure SDK that we used in our example application. 

We expect data-driven approaches, like ours, to further improve in the future as more data becomes available for  group detection. For the time being, our evaluation with small datasets showed that our approach performed better or as well as prior DL approaches \cite{Sanghvi2018-rssw, arxivearlier} across datasets. Additionally, our approach outperformed baselines on the Friends Meet dataset, a more general group detection benchmark than those considered in Sec. \ref{sec:experiments}. We believe that the success of our approach was in part due to its data-driven nature, as well as thanks to its structure. We predicted  affinity values for dyads instead of directly aiming to find groups, e.g., as in  \cite{arxivearlier}. This choice effectively increased the number of examples that \methodname\ had to learn from by a factor of 15 to 153, depending on the number of people in the scene, which helped avoid overfitting. Our results also suggest that \methodname\ effectively processed spatial data from a variable number of unordered social agents. Thus, our work reinforces the idea that symmetric functions are an effective mechanism for neural networks to deal with input sets \cite{qi2017pointnet,gupta2018social}.

\subsection{Improving Social Awareness in Practice}
We showed the applicability of our group detection approach in a human-robot interaction scenario. In this demonstration, our approach ran in real-time in a standard desktop machine. Because we considered the robot as one more social agent, our method detected not only human groups, but also human-robot groups. Furthermore, our approach was able to  detect when people joined or leaved the robot's conversation. The availability of this information opened up possibilities to improve the interaction with our robot photographer, e.g., by enabling it to better frame photos, or simply acknowledging the changes in its social context, which can help queue users of the photography system. 

Because \methodname\ is easily extensible, an interesting avenue for future work would be to augment the spatial features considered by our method. For example, other relevant features include: additional spatial features for the social agents, like body pose, which can signal social information \cite{joo2019towards}; features that describe the layout of the environment, which can affect F-Formations \cite{kendon1990}; or even information about the type of agent being analyzed. We believe that the latter type of information will be relevant for heterogeneous social interactions, e.g., interactions among people and virtual agents or robots. In these situations, agents may communicate attention and engagement through different non-verbal social signals. The relevance of these signals could potentially be captured by data-driven methods, like ours, to further improve group detection.  

\subsection{Limitations}

Our work is not without limitations. First, our method does not take advantage of the temporal correlation of spatial features captured by a situated sensor in the world. But this type of information could potentially improve group detection \cite{vazquez2015parallel}. Second, our method's runtime is also an important consideration. \methodname\ scales quadratically with the number of participants in the scene and the Dominant Sets \cite{hung2011detecting} clustering algorithm scales cubically. While this issue did not prevent us from performing real time group detection in our demonstration, it could be an issue if applied to more crowded settings. 
Third, our demonstration of our group detection approach was conducted in a laboratory setting. Further experiments are needed to validate the robustness of our approach in more dynamic scenarios.



\section{Conclusions}
We presented a novel approach for conversational group detection. Our method combined graph clustering with modern deep learning techniques to identify group interactions based on visual patterns of  spatial behavior. Under the challenging T=1 F1 metric, our method significantly outperformed or performed as well as previous methods in a variety of conversational group detection benchmarks. Additionally, we obtained good results under the GDSR metric in a more general group detection task, showing the generalization capabilities of our proposed approach. From an algorithmic point of view, clear improvements were derived from better affinity scores used for graph clustering in comparison to prior work. Additionally, the use of data-driven methods allowed our approach to cope with complex spatial patterns of behavior without ad-hoc steps to verify group interactions. These features made our approach robust and practical to be applied in a real human-robot interaction scenario. 




\begin{acks}
The Toyota Research Institute provided funds to assist with
this research.
This work is also partially funded by the National Science Foundation under Grant No. (IIS-1924802). Any opinions, findings, and conclusions or recommendations expressed in this material are those of the author(s) and do not necessarily reflect the views of any Toyota entity nor of the National Science Foundation.
\end{acks}

\bibliographystyle{ACM-Reference-Format}
\bibliography{bib}

\appendix

\section{Appendix}

Due to limited space, the evaluation described in the paper only presented overall results (averaged across five folds) for common group detection benchmarks. To supplement these results, Section \ref{Fold-by-Fold Results}  discusses group detection accuracy per fold. Afterwards, Section \ref{Synthetic Data Augmentation} discusses additional results when DANTE is trained with the addition of synthetic data, originally generated by Cristani et al. \cite{cristani2011social}. We hypothesized that synthetic data would help with group detection given the small size of existing datasets. The results did not demonstrate conclusively whether the synthetic augmentation improved \methodname.

\setlength{\tabcolsep}{0.3pt}
\begin{table}[b!p]
\centering
\caption{F1 results (T=1) for the Cocktail Party dataset.}
\label{tab:results-cparty}
\begin{tabular}{lccccccc}
\cline{1-7}
Model & Fold 1 & Fold 2 & Fold 3 & Fold 4 & Fold 5 & Overall  \\ \cline{1-7}
GComm & - & -  & - & - & - & 0.60  \\
GTCG & 0.31 & 0.46 & 0.09 & 0.17 & 0.43 & 0.29  \\
GCFF & 0.49 & 0.68 & 0.52 & 0.71 & 0.80 & 0.64 \\
DANTE & \textbf{0.73} & 0.80 & 0.56 & 0.72 & \textbf{0.83} & \textbf{0.73}  \\
DANTE+Synthetic & 0.70 & \textbf{0.83} & \textbf{0.59} & 0.64 & 0.79 & 0.71 \\
DANTE-NoContext & 0.60 & 0.68 & 0.33 & \textbf{0.75} & \textbf{0.83} & 0.64  \\
\cline{1-7}
\end{tabular}
\end{table}

\begin{table}[t!p]
\centering
\caption{F1 results (T=1) for the SALSA dataset.}
\label{tab:results-SALSA}
\begin{tabular}{lccccccc}
\cline{1-7}
Model & Fold 1 & Fold 2 & Fold 3 & Fold 4 & Fold 5 & Overall  \\ \cline{1-7}
GTCG & 0.39 & 0.44 & 0.41 & 0.47 & 0.50 & 0.44  \\
GCFF & 0.59 & 0.40 & 0.25 & 0.24 & 0.58 & 0.41 \\
DANTE & \textbf{0.76} & \textbf{0.57} & \textbf{0.70} & \textbf{0.50} & \textbf{0.69} & \textbf{0.65}  \\
DANTE+Synthetic & 0.74 & \textbf{0.57} & 0.59 & \textbf{0.50} & \textbf{0.69} & 0.62 \\
DANTE-NoContext & 0.70 & 0.41 & 0.63 & 0.46 & 0.64 & 0.57 \\
\cline{1-7}
\end{tabular}
\end{table}

\setlength{\tabcolsep}{0.3pt}
\begin{table}[t!p]
\centering
\caption{F1 results (T=1) for the Coffee Break dataset.}
\label{tab:results-cbreak}
\begin{tabular}{lccccccc}
\cline{1-7}
Model & Fold 1 & Fold 2 & Fold 3 & Fold 4 & Fold 5 & Overall   \\ \cline{1-8}
GComm & - & -  & - & - & - & 0.63  \\
GTCG & 0.48 & 0.32 & 0.51 & 0.48 & 0.58 & 0.48 \\
GCFF & \textbf{0.50} & 0.35 & 0.73 & 0.71 & 0.84 & 0.63  \\
DANTE & 0.39 & 0.40 & 0.74 & \textbf{0.77} & \textbf{0.89}   & 0.64 \\
DANTE+Synthetic & 0.48 & 0.36 & \textbf{0.77} & \textbf{0.77} & \textbf{0.89} & 0.65  \\
DANTE-NoContext & 0.46 & \textbf{0.43} & 0.76 & \textbf{0.77} & \textbf{0.89} & \textbf{0.66}\\
\cline{1-7}
\end{tabular}
\end{table}

\begin{table}[t!p]
\centering
\caption{F1 results (T=1) for the Friends Meet dataset.}
\label{tab:results-FM-F1}
\begin{tabular}{l c c c c c c c}
\cline{1-7}
Model & Fold 1 & Fold 2 & Fold 3 & Fold 4 & Fold 5 & Overall  \\ \cline{1-7}
GTCG & 0.71 & 0.70 & 0.60 & 0.62 & 0.67 & 0.65  \\
GCFF & 0.85 & 0.69 & 0.77 & \textbf{0.89} & 0.76 & 0.79\\
DANTE & \textbf{0.93} & 0.90 & 0.90 & 0.86 & 0.91 & 0.90  \\
DANTE+Synthetic & \textbf{0.93} & \textbf{0.91} & 0.90 & 0.87 & \textbf{0.94} & \textbf{0.91} \\
DANTE-NoContext & 0.91 & 0.89 & \textbf{0.92} & 0.88 & 0.91 & 0.90  \\
\cline{1-7}
\end{tabular}
\end{table}

\begin{table}[t!p]
\centering
\caption{GDSR results for the Friends Meet dataset.}
\label{tab:results-FM-GDSR}
\begin{tabular}{l c c c c c c c}
\cline{1-7}
Model & Fold 1 & Fold 2 & Fold 3 & Fold 4 & Fold 5 & Overall  \\ \cline{1-7}
GTCG &  0.95 & 0.89 & 0.91 & 0.86 & 0.87 & 0.90  \\
GCFF & 0.97 & 0.89 & 0.96 & 0.97 & 0.97 & 0.95 \\
DANTE & \textbf{0.99} & \textbf{0.99} & \textbf{0.98} & \textbf{0.99} & \textbf{0.99} & \textbf{0.99}  \\
DANTE+Synthetic & \textbf{0.99} & \textbf{0.99} & 0.96 & \textbf{0.99} & \textbf{0.99} & 0.98 \\
DANTE-NoContext & 0.98 & 0.98 & 0.97 & \textbf{0.99} & \textbf{0.99} & 0.98  \\
\cline{1-7}
\end{tabular}
\end{table}

\subsection{Fold-by-Fold Results} \label{Fold-by-Fold Results}
Recall that our main focus is to detect conversational groups, as annotated in the Cocktail Party \cite{zen2010space}, Coffee Break \cite{cristani2011social}, and SALSA \cite{SALSA} datasets. To test the generalization capabilities of our approach to a related but different group detection task, we also provide results for the Friends Meet dataset \cite{FriendsMeet}. Friends Meet was originally created for tracking diverse groups of people.

In Tables \ref{tab:results-cparty}, \ref{tab:results-cbreak}, \ref{tab:results-SALSA}, and  \ref{tab:results-FM-F1} we give a fold-by-fold breakdown of $T=1$ F1 results for the different datasets, as well as  overall (average) scores. Additionally, Table \ref{tab:results-FM-GDSR} gives a fold-by-fold breakdown of Group Detection Success Rate (GDSR) \cite{VASCON2017}  on the Friends Meet dataset. GDSR measures the percent of groups correctly identified, where a group is considered correct if 60\% of its members are included. It is easier than the challenging $T=1$ F1 metric which requires perfect group detection. Table \ref{tab:results-FM-GDSR} shows our method nearly saturates the GDSR metric both on a per-Fold and Overall basis.

The fold-level breakdown shows that DANTE often outperforms prior work with consistency and not simply on average. 
Note that noisy spatial features hurt prior work, but is especially harmful to our proposed method because of its dependency on data. This noise particularly affected our method in Fold 1 of the  Coffee Break dataset. 


\subsection{Synthetic Data Augmentation} \label{Synthetic Data Augmentation}

All the tables in the prior page report results from an experiment in which we augment \methodname's training data with synthetic examples from \cite{cristani2011social}. The synthetic data consists of 100 different situations created by psychologists. In each situation, some simulated people take part in F-Formations and others do not. Ground truth group as well as people positions and orientations are provided as part of this synthetic dataset.

Training DANTE with synthetic data lowered average group detection results by 2\% in the Cocktail Party dataset and 3\% in the SALSA dataset. In the Coffee Break dataset, the synthetic data slightly improved the results by 1\%. We attribute these mixed results to the differences in the amount of data that each benchmark provides and their quality.



\end{document}